\documentclass[conference]{IEEEtran}
\IEEEoverridecommandlockouts
\usepackage{cite}
\usepackage{amsmath,amssymb,amsfonts}
\usepackage{algorithmic}
\usepackage{graphicx}
\usepackage{textcomp}
\usepackage{xcolor}
\def\BibTeX{{\rm B\kern-.05em{\sc i\kern-.025em b}\kern-.08em
    T\kern-.1667em\lower.7ex\hbox{E}\kern-.125emX}}
\begin{document}

\title{KALMAN FILTER APPLIED TO A DIFFERENTIAL ROBOT\\

}

\author{
\IEEEauthorblockN{1\textsuperscript{st} Sendey Vera González}
\IEEEauthorblockA{\begin{minipage}{0.3\linewidth}\centering \textit{FACSISTEL} \\ \textit{Universidad Estatal Península de Santa Elena (UPSE)}\\
La Libertad, Ecuador \\
0000-0002-7144-0981 \end{minipage}}
\and
\IEEEauthorblockN{2\textsuperscript{nd} Luis Chuquimarca Jiménez}
\IEEEauthorblockA{\begin{minipage}{0.3\linewidth}\centering \textit{FACSISTEL} \\ \textit{Universidad Estatal Península de Santa Elena (UPSE)}\\
La Libertad, Ecuador \\
0000-0003-3296-4309 \end{minipage}}
\and
\IEEEauthorblockN{3\textsuperscript{rd} Douglas Plaza}
\IEEEauthorblockA{\begin{minipage}{0.3\linewidth}\centering \textit{FIEC} \\ \textit{Escuela Superior Politécnica del Litoral (ESPOL)}\\
Guayaquil, Ecuador \\
0000-0002-1399-0627 \end{minipage}}
}

\maketitle

\makeatletter
\def\footnoterule{\kern-3\p@
  \hrule \@width 2in \kern 2.6\p@} 
\makeatother
\newcommand{\copyrightnotice}[1]{{%
  \renewcommand{\thefootnote}{}
  \footnotetext[0]{#1}%
}}
\copyrightnotice{979-8-3503-1576-9/23/\$31.00 ©2023 IEEE \newline © 2024 IEEE. Personal use of this material is permitted. Permission from IEEE must be obtained for all other uses, in any current or future media, including reprinting/republishing this material for advertising or promotional purposes, creating new collective works, for resale or redistribution to servers or lists, or reuse of any copyrighted component of this work in other works.  
\newline
DOI: https://doi.org/10.1109/10291441}

\begin{abstract}
This document presents the study of the problem of location and trajectory that a robot must follow. It focuses on applying the Kalman filter to achieve location and trajectory estimation in an autonomous mobile differential robot. The experimental data was carried out through tests obtained with the help of two incremental encoders that are part of the construction of the differential robot. The data transmission is carried out from a PC where the control is carried out with the Matlab/Simulink software. The results are expressed in graphs showing the path followed by the robot using PI control, the estimator of the Kalman filter in a real system.
\end{abstract}


\section{Introduction}
 This document presents a study on the problem of the trajectory of a robot in a confined area, the basic Proportional–Integral–Derivative (PID) control will be applied for the movement control of the mobile robot, and later an analysis of the application of the use of the Kalman filter\cite{b17}.
The test scenario is carried out using a real differential robot, implemented during the development of this work, to apply the FK within a noisy system estimating position data and obtaining the minimum amount of errors along the robot's trajectory\cite{b1,b2}.

\section{MECHANICAL DESIGN}
It is proposed to identify the parameters of the plant through the use of a direct and inverse kinematic model for a differential robot, with the application of PI control and the Kalman Filter to correct the robot's trajectory\cite{b3,b4,b5}.

To check that the plant and control system is operating, a two-wheeled mobile robot with micromotors is built with a microcontroller, encoder sensors, and a USB connection for communication with the PC.

The Kalman filter observer will help correct the trajectory of the mobile robot, estimating its next position. Employing a classic control, the micromotors will be acted on, and these will act on the wheels to execute the trajectory. Using the encoder will allow calculating the robot's speed and estimated position from an initial reference. The software for this project is MATLAB for mathematical modeling, calculations, and estimation, and ARDUINO, the embedded system, is a data acquisition and control execution system\cite{b6}.

The purpose of the differential robot is to move autonomously in a previously planned trajectory within a frame of reference. The movements made by the robot are consequences of the turning speeds of the motors. Each motor rotates according to the polarity and voltage applied on each terminal. The PWM signal allows control to increase or decrease the rotation speed of each DC motor, and the rotation speed of each motor is measured in angular speed, whose units are revolutions per second.

For the robot to turn on its axis to the right or the left, a linear velocity v of the robot equal to zero and an angular velocity w of the robot different from zero must be applied. In practice, this means that both wheels rotate in the same direction. Same angular velocity but in the opposite direction.

 \begin{figure}[htbp]
\centerline{\includegraphics{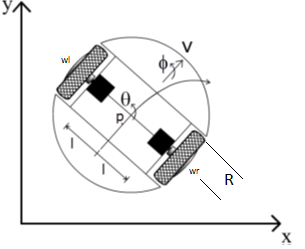}}
\caption{Angular speed wl and wr, angular and linear direction of the robot.}
\label{fig1}
\end{figure} 
  
\subsection{Kinematic models of the robot}
The kinematic behavior of the robot is due to the application of the direct and inverse models of the kinematics for a two-wheeled differential robot, where v,w are the velocities of the robot and Vr and Vl are the linear velocities of the robot with the left wheel and right respectively for each wheel 
equation \ref{eq1}, and equation \ref{eq2} respectively (see Figure \ref{fig1}).

\begin{equation}
 v= R\frac{\left ( Vr+Vl \right )}{2}
 \label{eq1}
\end{equation}

\begin{equation}
 w= R\frac{\left ( Vr-Vl \right )}{l}
 \label{eq2}
\end{equation}

Therefore, to define the position and orientation of the robot, the respective equations \ref{eq3}, \ref{eq4}, and \ref{eq5} are integrated.
\begin{equation}
\dot{x}= v cos \theta 
 \label{eq3}
\end{equation}

\begin{equation}
\dot{y}= v sin \theta 
 \label{eq4}
\end{equation}

\begin{equation}
\dot{\phi }=w
 \label{eq5}
\end{equation}

A rotation matrix is proposed since the robot is not always aligned with the global axis. The matrix equations are obtained from equations \ref{eq6} and \ref{eq7}.

\begin{equation}
\begin{bmatrix}
\dot{x}\\ 
\dot{y}\\ 
\dot{\phi }
\end{bmatrix}= \begin{bmatrix}
\cos \theta  & 0\\ 
\sin \theta  & 0\\ 
0 & 1
\end{bmatrix}\begin{bmatrix}
v\\ w
\end{bmatrix}
 \label{eq6}
\end{equation}

\begin{equation}
\begin{bmatrix}
\dot{x}\\ 
\dot{y}\\ 
\dot{\phi }
\end{bmatrix}= \begin{bmatrix}
\cos \theta  & 0\\ 
\sin \theta  & 0\\ 
0 & 1
\end{bmatrix}\begin{bmatrix}
R\frac{\left ( v_{r}+ v_{l}\right )}{2}\\R\frac{\left ( v_{r}-v_{l} \right )}{l}
\end{bmatrix}
 \label{eq7}
\end{equation}

The matrix equation \ref{eq8} represents the changes of variables (x,y, phi) as a function of the angular speeds of each wheel.

\begin{equation}
\begin{bmatrix}
\dot{x}\\ 
\dot{y}\\ 
\dot{\phi }
\end{bmatrix}= \begin{bmatrix}
R\frac{\cos \theta }{2} & R\frac{\cos \theta }{2} \\ 
R\frac{\sin \theta }{2} & R\frac{\sin \theta }{2} \\ 
\frac{R}{l} & -\frac{R}{l}
\end{bmatrix}\begin{bmatrix}
w_{r}\\w_{l}
\end{bmatrix}
 \label{eq8}
\end{equation}

In this system, the input is the linear and angular velocity of the robot. From the inverse kinematics, the angular velocities of each of the wheels will be obtained.

Sensing the angular velocities of the wheels and from the direct kinematics in equation \ref{eq9} and \ref{eq10}, the robot's real linear and angular velocity is calculated\cite{b7}.
\begin{equation}
\begin{bmatrix}
w_{r}\\ 
w_{l}
\end{bmatrix}= \begin{bmatrix}
\frac{1}{R} & \frac{l}{R}\\\frac{1}{R} 
 &-\frac{l}{R} 
\end{bmatrix}\begin{bmatrix}
v\\ \phi\
\end{bmatrix}
 \label{eq9}
\end{equation}

\begin{equation}
\begin{bmatrix}
v\\\ 
\phi 
\end{bmatrix}= \begin{bmatrix}
\frac{R}{2} & \frac{R}{2}\\\frac{R}{R}l 
 &-\frac{R}{2} l
\end{bmatrix}\begin{bmatrix}
w_{r}\\ 
w_{l}
\end{bmatrix}
 \label{eq10}
\end{equation}

\begin{figure*}[htb]
    \centering
    \includegraphics[width=1\textwidth]{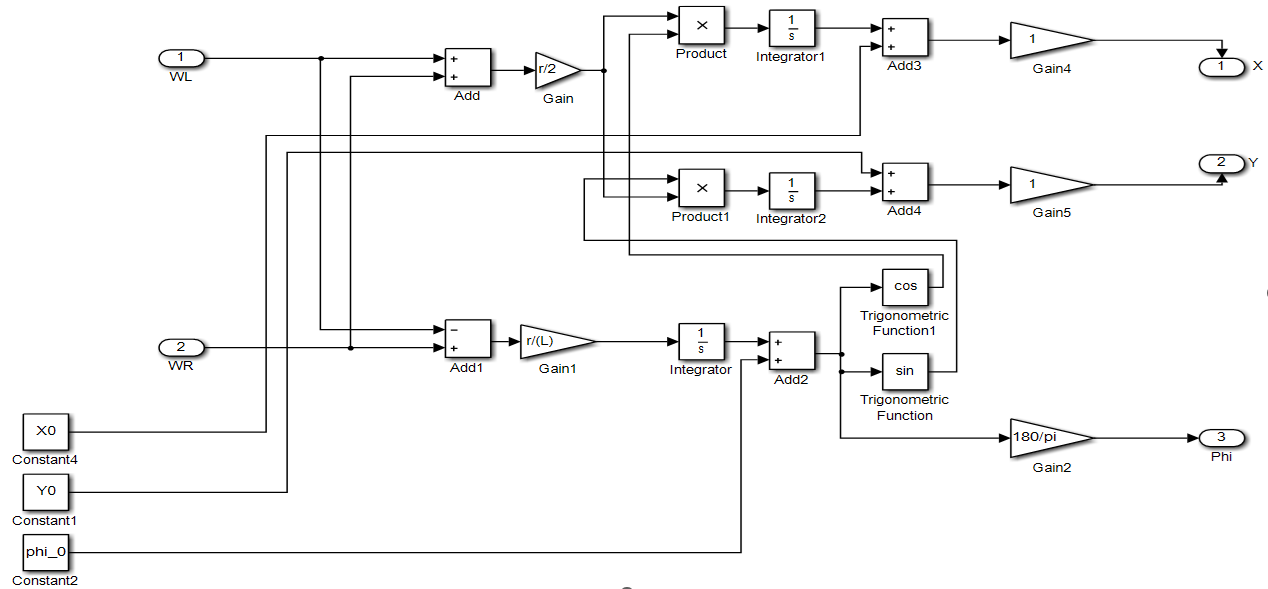}
    \caption{Direct kinematic model diagram of the robot.}
    \label{fig2}
\end{figure*}

\begin{figure}[h]
    \begin{center}
    \includegraphics[scale=0.6, angle = 0]{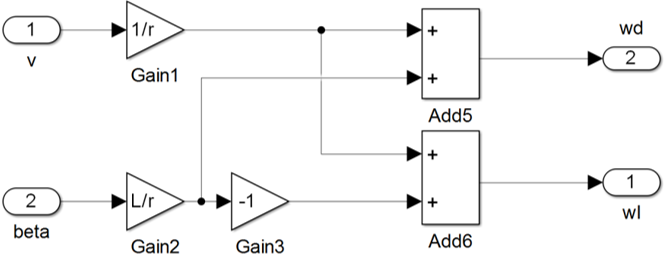}
    \caption{Inverse kinematic model diagram of the robot.} \label{fig3}
    \end{center}
\end{figure}

Next, the schemes made in Matlab-Simulink of the equations proposed for each model (see Figure \ref{fig2}, and \ref{fig3}).

\section{ELECTRONICS}
\begin{figure*}[htb]
    \centering
    \includegraphics[width=0.6\textwidth]{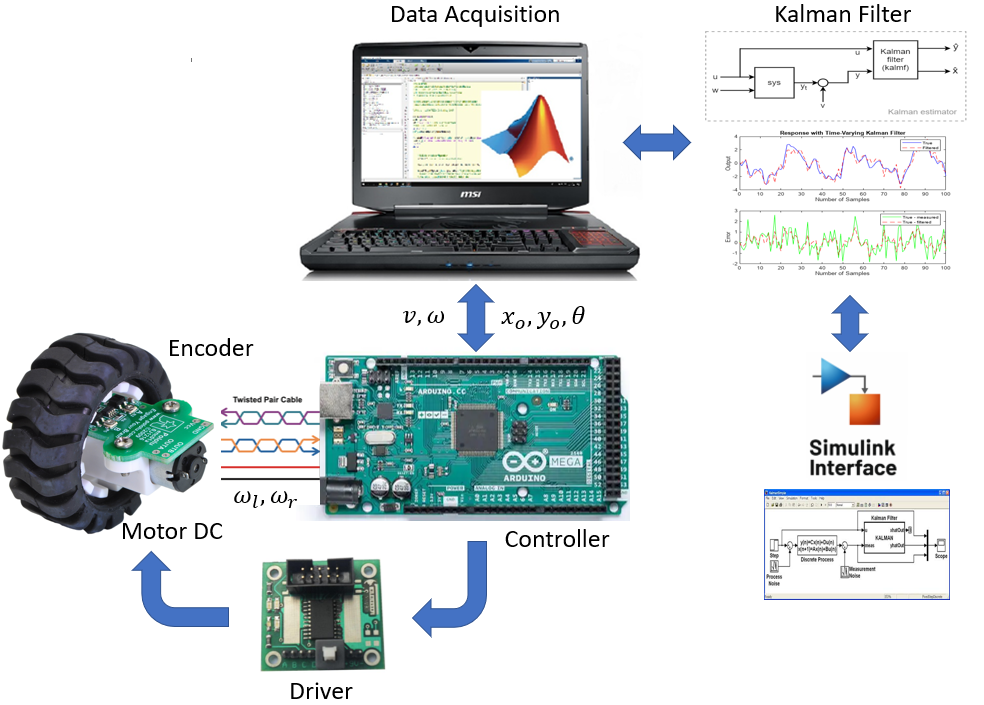}
    \caption{The interconnection of elements between the plant and the computer with the Matlab-Simulink interface.}
    \label{fig4}
\end{figure*}

 The interface for the robot's communication with Matlab Simulink uses the MathWorks ArduinoIO package.  This package allows relatively transparent and real-time communication between Simulink and Arduino\cite{b8,b9,b10} (see Figure \ref{fig4}).
   
 The firmware adioes.pde from the ArduinoIO package must be loaded on the Arduino board, and then from Matlab-Simulink send and receive information. In this way, the Arduino card performs the functions of a data acquisition card and generates the output to control the micromotor.
   
\subsection{SOFWARE}

\begin{figure*}[htb]
    \centering
    \includegraphics[width=1\textwidth]{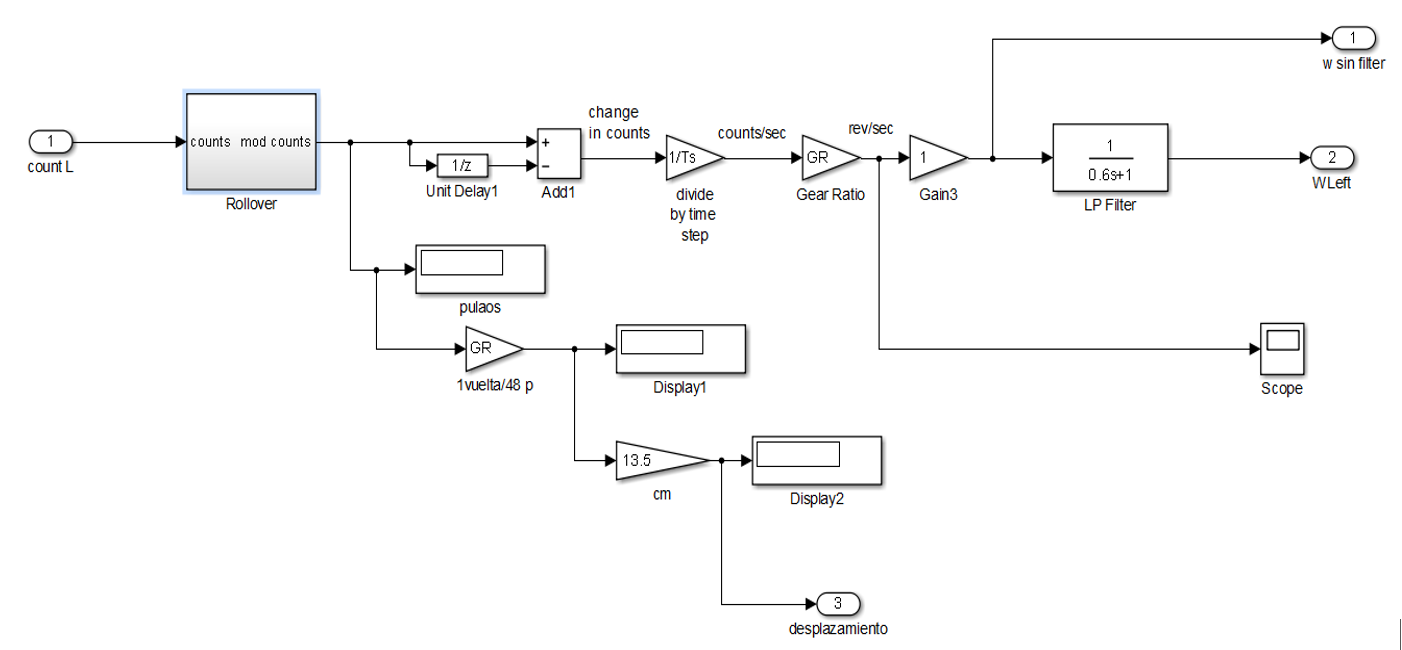}
    \caption{The graphical programming made in Matlab/Simulink.}
    \label{fig5}
\end{figure*}

 \begin{figure}[htbp]
\centerline{\includegraphics{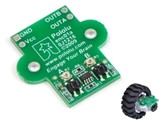}}
\caption{Encoder for Pololu Wheel 42x19mm.}
\label{fig6}
\end{figure}

Figure \ref{fig5} shows the graphical programming made in Matlab/Simulink. In the upper part of the graph, it is used not to exceed the memory limit of the temporary register where the information is stored and to obtain the direction in which the micromotor rotates\cite{b11,b12,b13}. Then count each pulse and obtain a sample of changes every 2 ms.
 
The tests with the oscilloscope and video camera to record the lap numbers result in 10 laps in one second at a power of 20 PWM.

The calculation of distance in terms of the linear speed of the robot is from 0.1 m/s to 4.8 rad/s, that is, 0.76 turns in a second. For example, for 76 turns, 10,944 meters are traveled in a straight line.

Figure \ref{fig6} shows the type of encoder used in the differential robot. According to the manufacturer, it allows us to determine the direction of rotation and provide the four counts for each tooth\cite{b14}. A resolution of 48 counts is obtained for each revolution of the wheel.

\subsection{CONTROL SYSTEM}
\begin{figure}[htbp]
\includegraphics[width=7cm, height=4cm]{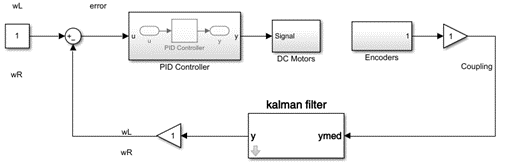}
\caption{PI control system for angular speed.}
\label{fig7}
\end{figure}

As mentioned above, the angular speed of the left and right micromotor are the signals to be controlled, for which a PI feedback loop is used, with the encoder reading and the angular speed reference for a robot trajectory\cite{b15}.
The change in speed of each micromotor causes a change of direction of the robot's trajectory. Figure \ref{fig7} shows the feedback scheme.

\begin{figure}[htbp]
\includegraphics[width=8cm, height=5cm]{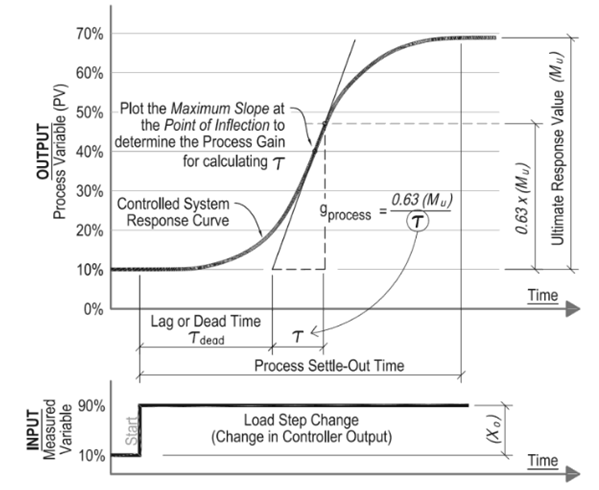}
\caption{Ziegler-Nichols reaction curve.}
\label{fig8}
\end{figure}

A low pass filter is used to smooth the speed output waveform obtained by the encoder.
The graphical method is applied to calculate the PID tuning of the Ziegler-Nichols reaction curve to obtain the control parameters (see Figure \ref{fig8})\cite{b16}.

 Therefore the value recorded for the PID is kp=0.375, Ki=1 and Kd=0 
 \subsection{Kalman filter }  
 The Kalman filter is essentially a set of mathematical equations that implement a predictor-corrector estimator that is optimal in the sense that it minimizes the covariance of the estimated error. The great advantage of the Kalman filter is its relative "simpleness" and robustness since it can work considerably well in many situations.

\begin{figure}[htbp]
\includegraphics[width=8.8cm, height=5cm]{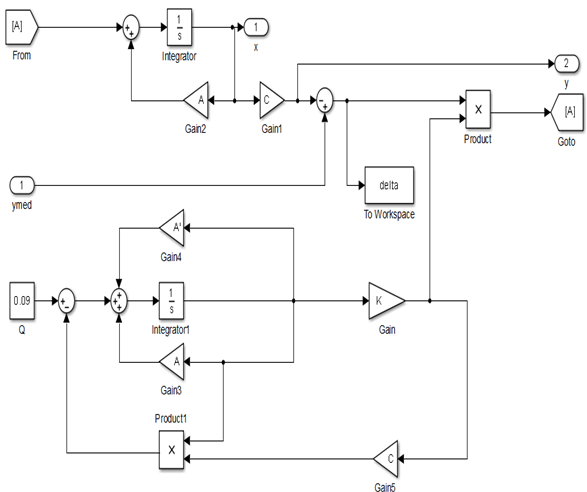}
\caption{FK estimator algorithm.}
\label{fig9}
\end{figure}

Figure \ref{fig9}  shows the graphic programming in Matlab-Simulink of the Kalman filter for one variable.

The meaning of the variances Q and W is as follows: if we are sure that the noise of the process little influences the equations of state, the initial value of Q is small. On the contrary, if unsure, we must choose Q's most significant initial value.

In the same way, for the choice of the value of the variance W, if we are sure that the measurement carried out is a little degraded by the measurement noise, then the small value of W is chosen. On the contrary, if we are unsure of the certainty of a sensor measurement, then the measurement is degraded, and the variance value W is large.

\section*{RESULTS}
\begin{figure}[h]
\includegraphics[width=8cm, height=5cm]{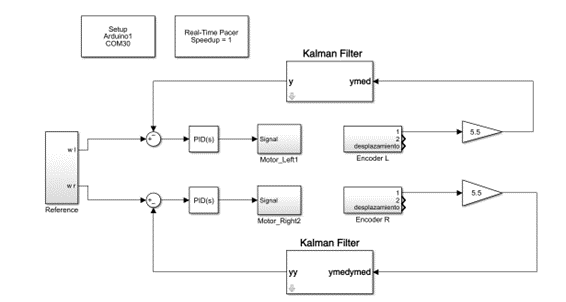}
\caption {Summary algorithm of the PID control plus FK of the differential robot.}
\label{fig10}
\end{figure}

The test design consists of executing the algorithm in Matlab/Simulink of Figure \ref{fig10} to demonstrate the performance of applying the Kalman filter equations to the linear system. The variables involved in this system are the measurements collected by the optical sensor, which comprises the encoders, and the noise involved as the estimation output of the filter on each wheel.

\subsection{Activation of the robot for the hexagon path with PID + FK control.} 

\begin{figure}[h]
\includegraphics[width=8cm, height=5cm]{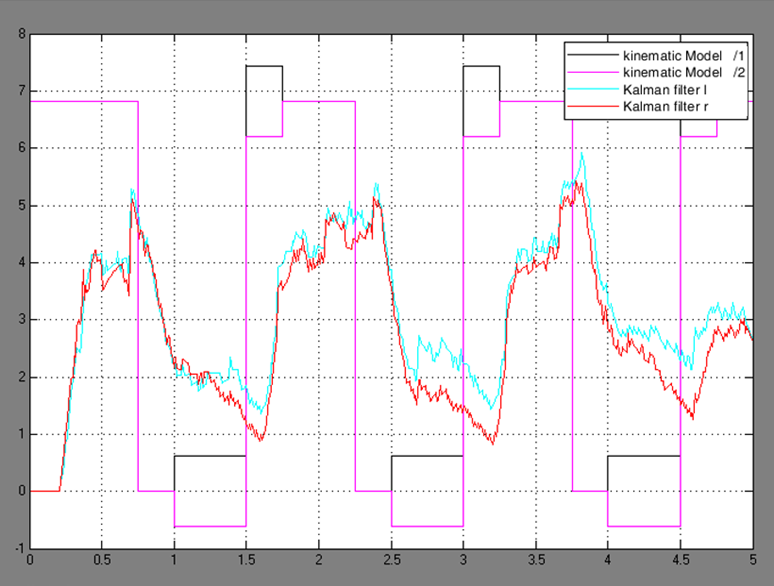}
\caption{Reference angular speed signal and angular speed reading in each micro motor.}
\label{fig11}
\end{figure}

\begin{figure}[htbp]
\includegraphics[width=8cm, height=5cm]{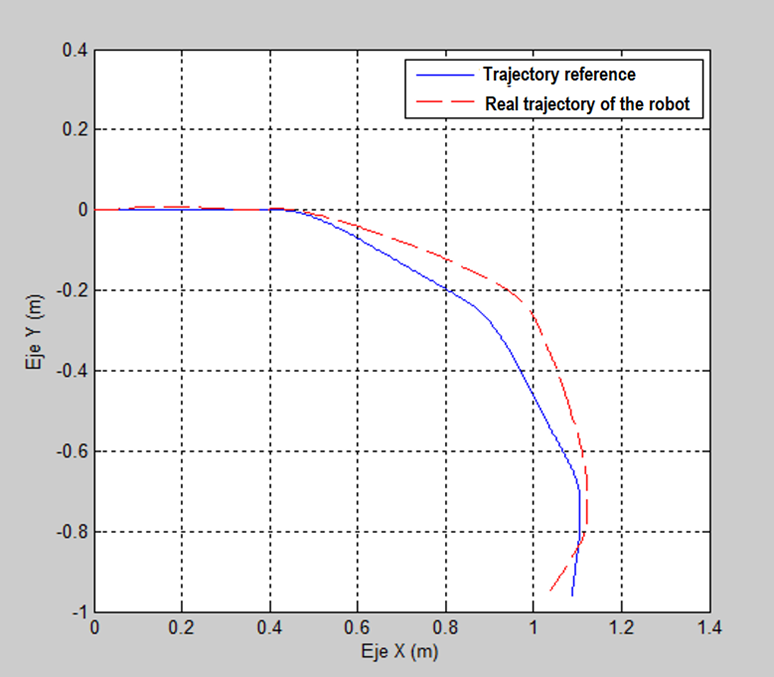}
\caption{Reference path and path recorded by the robot.}
\label{fig12}
\end{figure}

Then the performance of the Kalman filter to estimate the angular speed of the robot's micromotors can be seen by changing the reference speeds.
The angular velocity changes applied to the wheels produce the different trajectories of the robot. In this test, the trajectory had three turns, three-speed changes angle for each micro motor, and the test time lasted 5 seconds. The trajectory begins with the advance in a straight line until 0.7 seconds with a response of 4 rad/s. After that, there is a motor shutdown to start with a turn to the left, the turn time is 0.7 seconds, but it should be noted that the turn has two parts, a low speed. A fast speed, this event in the motors causes the effect of changes in the wheels first looking to stop in less than 0.25 seconds, to start then turning with different speeds of 0.6 and -- 0.4 rad/s, after this, it increases very quickly until reaching 7.5 rad/s and 6.2 rad/s with which the motor rotates until reaching the reference, this is done in 0.25 seconds, and the process is repeated. The trajectory was remarkably acceptable because the robot followed the same trajectory for almost the entire route (see Figure \ref{fig11} and \ref{fig12}).

\begin{figure}[htbp]
\centerline{\includegraphics{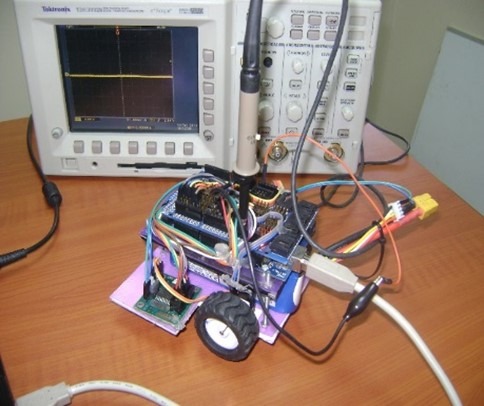}}
\caption{Robot Differential MC.}
\label{fig13}
\end{figure}

The results of the experimental tests carried out on the differential robot are as follows (see Figure \ref{fig13}):

1.- .The graphs of the real trajectory and the proposed trajectory of the robot use the direct kinematics model. The calibration of the real measurements is due to the conditions of adequate power to overcome the friction between the wheels and the ground. When considering the robot's movement on a plane (x,y), this minimum power was obtained from the experimental test, where 30 PWM is applied to overcome inertia and whose wheel speed changes is 1 rad/s.

2.- The results show each wheel's proposed and real angular speeds. This is due to applying the inverse kinematic model to trace the route. The input variables to the model are the linear and angular velocities of the robot, and the model outputs are the angular velocities of each wheel. These angular velocity readings applied to the real robot showed a linear change of 1 rad/s for every 30 PWM for a range between 1 and 6 rad/s, and after that, there is an exponential change of 2 rad/s for every 30 PWM. registered.

3.- The PID controller with the following parameters Kp=0.479 and Ki =1 applied on the plant gives us results of convergence towards the angular speed in maximum times of up to 3 seconds stabilization time, and depending on the speed that has been applied at the input the fastest response of stabilization time is less than 0.5 seconds, there is a small lag of the left wheel possibly due to motor wear or encoder calibration, even so, the PID control does the control causing a small oscillation in the (test 2) without FPB but with a faster response over control and instead than the 2-second stabilization as a result of PID control with an FPB.

4.- On the results obtained by applying the Kalman filter with the parameters of A=-0.0001 C=1; K=1000 and the PID control of information from a noisy optical encoder and using an estimator allows for reducing the amplitude of the noisy signal component at the output of the speed control system. It does not present damped oscillations and has a fast response with a Kalman filter as a state estimator.

\section{CONCLUSIONS}
The feedback information that is used from the plant is the angular speed, then the encoder option is adopted to obtain the angular speed measurement of the micromotor, this speed of each motor is the cause of the robot movement.
 
The resolution of each incremental encoder is 48 pulses per revolution in 4X; according to the manufacturer, the angle change for each pulse in each wheel is 7.5 degrees, a relatively high amount for changes of small turns. Therefore the angular velocities in each wheel must be high to obtain more reliable readings. It was decided to work with values between 5 and 7 rad/s. On the other hand, it was noted that a signal range between -20 to +20 PWM did not generate enough power in the micromotor to move the robot; therefore, the angular speed control system is considered linear in an angular speed restriction between +2 and +7 rad/s that generate a power signal of at least +30 PWM.

The control model used for the robot movement is based on the difference between the angular speed of the reading obtained from the encoders placed on the robot wheels and the angular speed applied to the motor on both wheels. To reduce this error, a PI control and the Kalman filter were applied to the system. The trajectory traveled by the differential robot shows that by using the direct and inverse kinematics model, it is possible to represent the desired trajectory and the real route that the robot has made. Depending on tests, these trajectories are presented in a 2D map within a frame of reference up to 100 cm x 100 cm. The direct kinematic model is used to obtain the robot's location or position of (x, y), given the angular speed of each micromotor. And the inverse kinematics model is used to obtain the angular speed required in each micromotor given the differential robot's proposed linear and angular speed.
       
To tune the PI control and correct the error with respect to the reference, the parameters Kp and Ki were calculated twice. The first time a low pass filter was applied to the encoder output, it was observed that in the graphic results, the angular speed reading is not immediate and is delayed by 3 seconds in reaching the reference speed, later a new calculation is made, applying the Kalman filter to the direct output of the incremental encoder without FPB. It was obtained that the reference speed is reached faster than the previous one, this is in just 0.5 seconds according to the experimental tests, which means that when using the FK without FPB in the system, the response time is decreased to reach the reference speed.

For the Kalman filter design, the required and suggested parameters are calculated; these were obtained by the characteristics of the encoder which has a common Gaussian noise. According to the bibliography consulted for these types of the encoder, acceptable values of P = -0.0001 and for Q = 10000, which were the values suggested and approximate to those calculated by the experimental results of motor speed with load and without load, with quite reliable results according to the tests of the micromotor - encoder.

Three experimental tests were designed to verify the operation of the PID control plus the FK for the angular speed of the wheels of the micromotors. These speeds serve to fulfill a linear, circular path and a combination of both. In all cases, there was the tracking of the trajectory with small errors when following the desired trajectory. In the case of not achieving the desired trajectory, this is due to the possible inaction or late reaction in a motor, the real trajectory of the differential robot of According to the experimental tests carried out with the same control parameters, it was noted that one of the micromotors reacts more slowly than the other.

\end{document}